\documentclass{article}
\usepackage{spconf,amsmath,graphicx}
\usepackage{subfigure}
\usepackage{url}

\title{Online PCB Defect Detector On a New PCB Defect Dataset}
\name{Sanli Tang, Fan He, Xiaolin Huang, Jie Yang\thanks{Corresponding author: Jie Yang; e-mail: jieyang@sjtu.edu.cn}}
\address{Institute of Image Processing and Pattern Recognition, Shanghai Jiao Tong University, China}
\begin{document}
%
\maketitle
\begin{abstract}

Previous works for PCB defect detection based on image difference and image processing techniques have already achieved promising performance. However, they sometimes fall short because of the unaccounted defect patterns or over-sensitivity about some hyper-parameters. In this work, we design a deep model that accurately detects PCB defects from an input pair of a detect-free template and a defective tested image. A novel group pyramid pooling module is proposed to efficiently extract features of a large range of resolutions, which are merged by group to predict PCB defect of corresponding scales.
To train the deep model, a dataset is established, namely DeepPCB, which contains 1,500 image pairs with annotations including positions of 6 common types of PCB defects. Experiment results validate the effectiveness and efficiency of the proposed model by achieving $98.6\%$ mAP @ 62 FPS on DeepPCB dataset. This dataset is now available at: \url{https://github.com/tangsanli5201/DeepPCB}.
\end{abstract}
\section{Introduction}
\label{sec:intro}
PCB manufacturing has drawn more and more attention as the rapid development of the consumer electronic products. Generally, manual visual inspection is one of the biggest expense in PCB manufacturing. As popular and non-contact methods, recent researches \cite{PCBHybrid, PCBMorphology, PCBsorting} propose to process both a defect-free template and a defective tested image to localise and classify defects on the tested image. However, those image-based PCB defect detecting algorithms are often challenged by lacking of sufficient data with elaborated annotations to validate their effectiveness, which also prevents the researches from training an advanced detector, e.g. the neural network.

Earlier works on PCB defect detection focus on wavelet-based algorithms \cite{PCBWavelet1, PCBWavelet2, PCBWavelet3, PCBWavelet4, PCBWavelet5}, which decreases the computation time compared to those based on image difference operation. Recently, \cite{PCBMorphology} develops a hybrid algorithm to detect PCB defects by using morphological segmentation and simple image processing technique. \cite{PCBCVS} incorporates proper image registration to solve the alignment problem, which however consumes more processing time.  \cite{PCBKNN} proposes to use image subtraction and KNN classifier, which greatly improves the performance. \cite{PCBHoles} provides a technique to separate holes and find their centroids, which are then used with the connected components to identify the defects. \cite{PCBDataMining} uses data mining approach to classify PCB defects based on a private dataset. However, those algorithms relying on image difference and logic inference sometimes fail due to: (a) the complicated or unaccounted defect patterns; (b) irregular image distortion and offset between the template and tested image pair; (c) over-sensitivity of hyper-parameters, e.g. the kernel size of erosion or dilation operation.

\begin{figure} [t] 
\centering
\begin{minipage}{3.8cm}
\includegraphics[width=3.8cm]{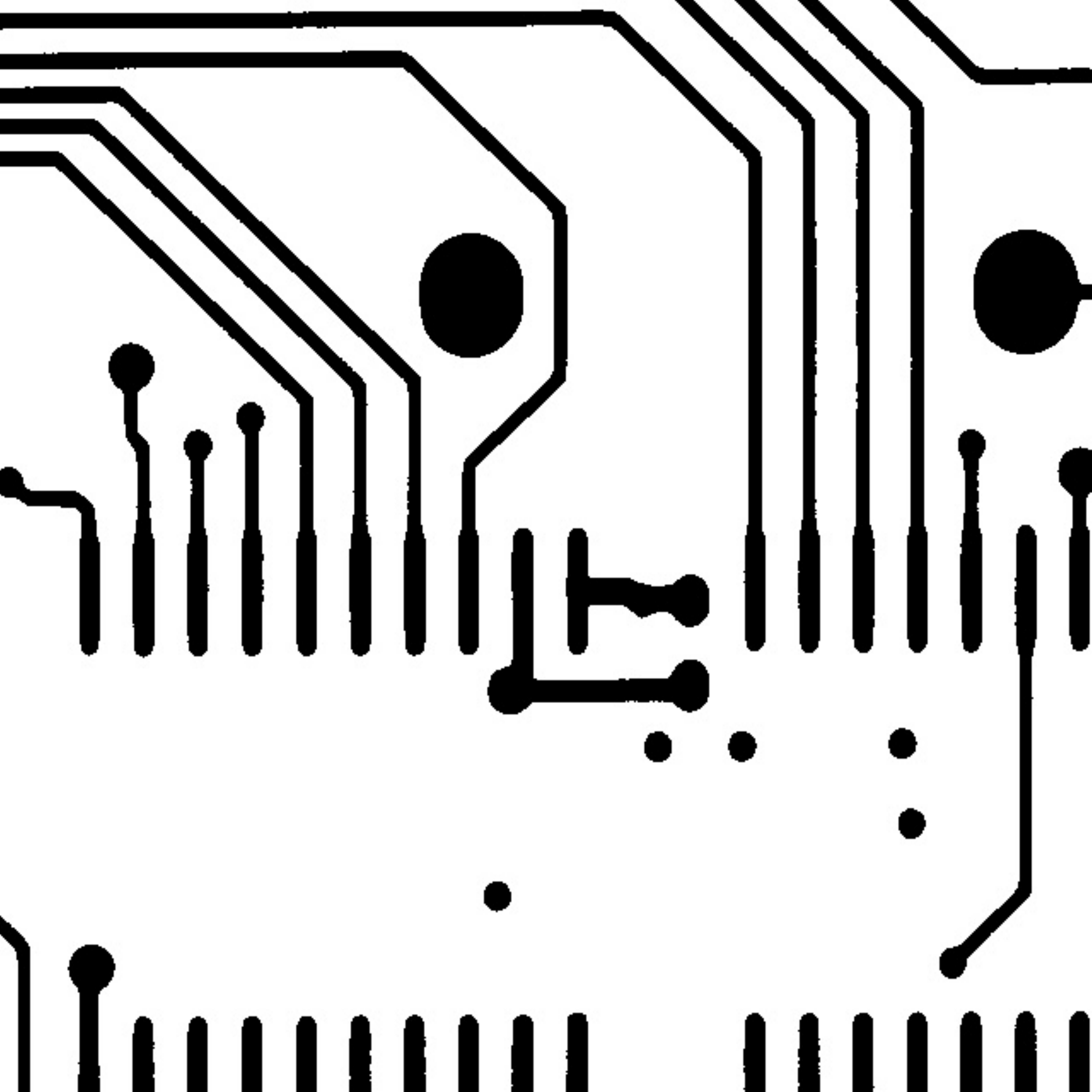}
\centerline{(a)}
\end{minipage}
\hspace{0.5cm}
\begin{minipage}{3.8cm}
\includegraphics[width=3.8cm]{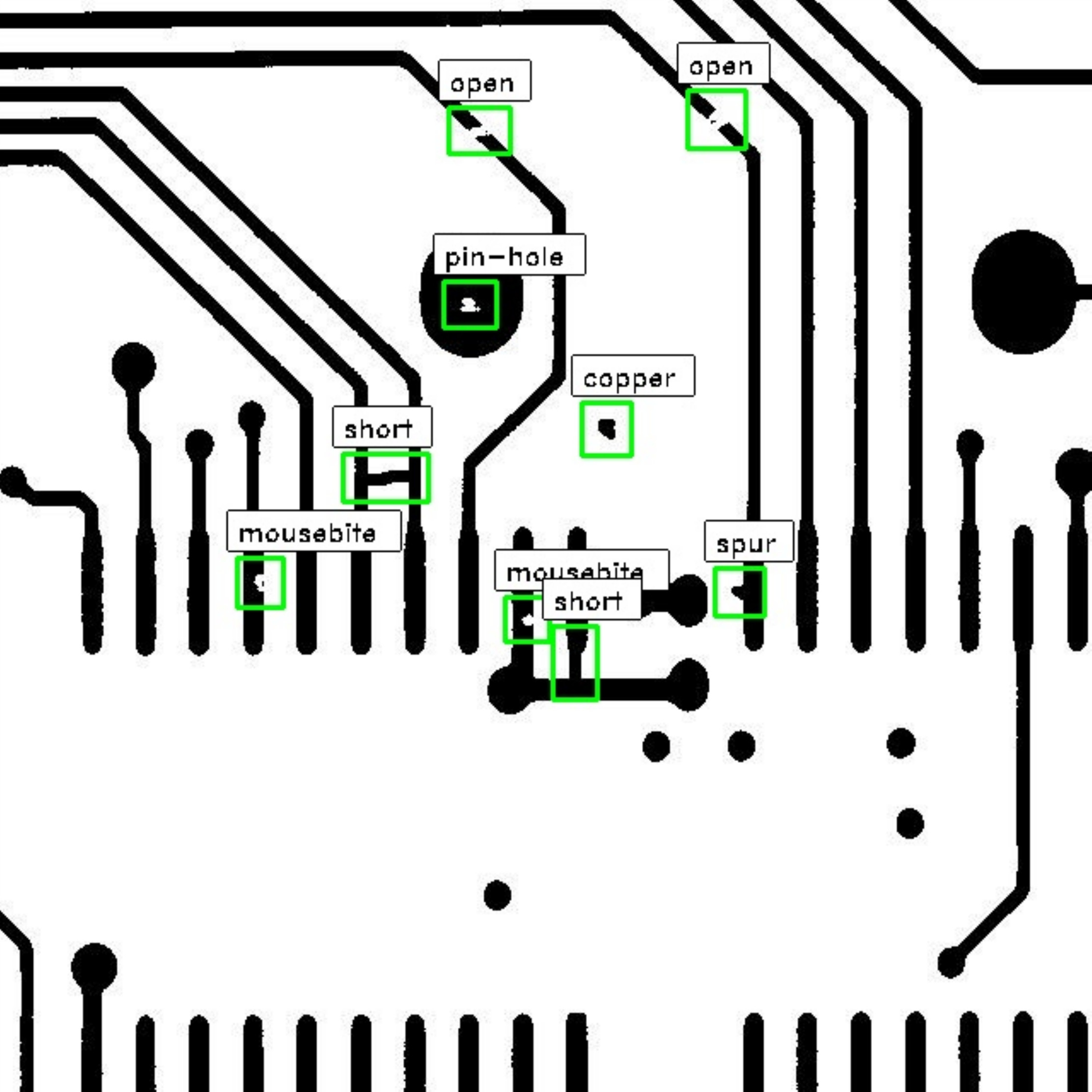}
\centerline{(b)}
\end{minipage}
\caption{This figure shows the pair of (a) a defect-free template image and (b) a defective tested image with annotations of the positions and types of PCB defects in the DeepPCB dataset.}
\label{DatasetIllustration}
\vspace{-0.1cm}
\end{figure}

Recently, deep neural network has shown its strong generalization aiblity on object detection task \cite{SSD, Faster, DSSD, Focalloss}. PCB defect detection is essentially a kind of object detection task with two slight differences: (1) object as the PCB defect and (2) the pair-wise input including a defect-free template image and a defective tested image. As for object detection models based on convolutional networks, \cite{Faster} proposes a two stage object detection framework, in which a Region Proposal Network (RPN) is first deployed to generate high-quality region proposals that are used by Fast R-CNN for detection. \cite{YOLO} proposes to bypass the region proposal stage as in \cite{Faster} and directly make prediction of the positions and classes for objects in the input image, which slightly sacrifices the precision but runs about 7 time faster than \cite{Faster}. \cite{SSD} makes prediction from multiple feature maps of different resolutions to deal with objects of various sizes, which achieves impressive performance while keeping low inference time. Besides, \cite{FPN} provides a feature pyramid network to merge features from different resolutions in a bottom-up manner, which is also a promising structure to detect objects in various scale.


To train an advanced deep model for PCB defect detection, in this work, we first set up a dataset, namely DeepPCB,  which includes 1,500 pairs of template and tested images with annotations of position and class of 6 types of PCB defects. To the best of our knowledge, this is the first public dataset for PCB defect detection. As illustrated in Fig \ref{DatasetIllustration}, this dataset enjoys several advantages. (1) Alignment - the template and tested images are aligned by template matching method, which reduces lots of effort for image preprocessing.
(2) Availability - DeepPCB will be public for research purpose. 
We believe that this dataset will be greatly beneficial to the research in PCB defect detection.

PCB defect detectors based on the advanced deep models usually are faced with the dilemma of the accuracy and the efficiency. High accuracy requires much deeper models with tens and hundreds of layers to obtain higher level features in larger respective field, while high efficiency needs much fewer parameters as well as the less depth structure. To reduce the contradiction, we propose a novel module, namely group pyramid pooling (GPP), which merges features in various resolutions from grouped pooling and up sampling. Each group in GPP carries both local and much larger range of context information and takes responsibility for predicting PCB defects in corresponding scales.


This paper makes three contributions. (1) We propose a PCB defect detection network based on the novel module: group pyramid pooling (GPP), which increases the model's ability of detecting PCB defects in various scales. (2) The first dataset for PCB defect detection is established, which includes 1,500 aligned image pairs with precise annotations. (3) Extensive experiments are carried out to validate the effectiveness and efficiency of the proposed 
model and the usefulness of the DeepPCB dataset.

\section{The DeepPCB Dataset}
\label{sec:pagestyle}

We contribute DeepPCB to the community, which contains 1,500 PCB image pairs covering six types of PCB defect. Each pair consists a 640 x 640 defect-free template image and a defective tested image. We separate 1,000 image pairs as training set and the remaining 500 image pairs as test set.

\begin{figure}[h]
\centering
\includegraphics[width=8.5cm]{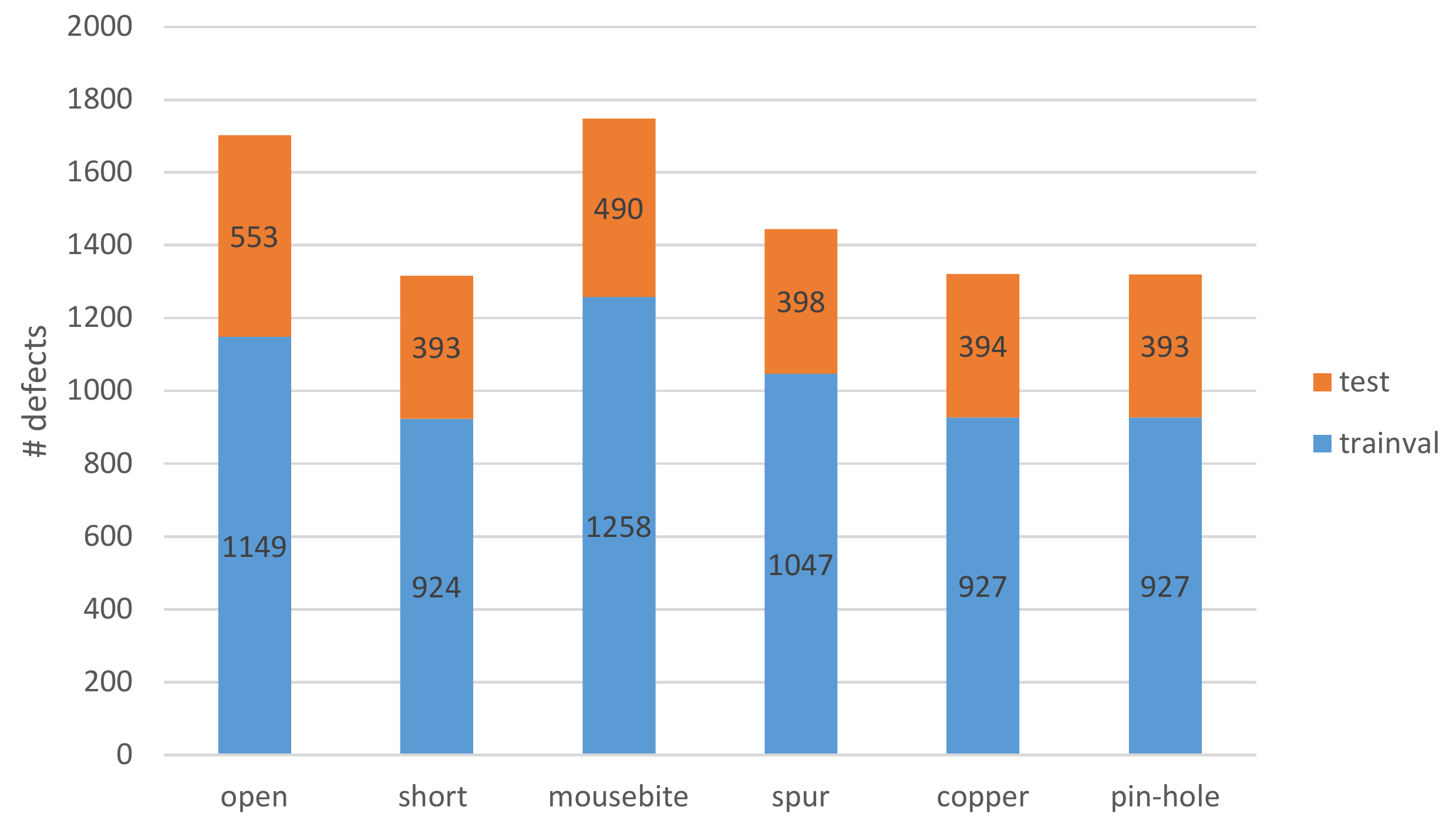}
\caption{Defect number of the 6 categories in DeepPCB train/validation and test set.}
\label{FigDefectCount}
\end{figure}

\subsection{Image Collection}
Following the common industrial settings, all the images in this dataset is obtained from a linear scan CCD in resolution around 48 pixels per 1 millimetre. The defect-free template image is manually checked and cleaned from a sampled image in the above manner. The original size of the template and tested image is around 16k x 16k pixels. Then they are clipped into sub-images with size of 640 x 640 and aligned through template matching techniques by reducing the translation and rotation offset between the image pairs. Next, a threshold is carefully selected to employ binarization to avoid illumination disturbance. Although there are different prepocessing methods according to the specific PCB defect detecting algorithm, the image registration and thresholding are common techniques for high-accuracy PCB defect localization and classification \cite{Survey}.

\vspace{-0.1cm}
\subsection{Image Annotation}
We use the axis-aligned bounding box with a class ID for each defect in the tested images. As illustrated in Fig. \ref{DatasetIllustration}, we annotate six common types of PCB defects: open, short, mousebite, spur, pin hole and spurious copper. Since there are only a few defects in the real tested image, we manually argument some artificial defects on each tested image according to the PCB defect patterns \cite{Survey1996}, which leads to around 3 to 12 defects in each 640 x 640 image. The number of PCB defects is shown in Fig. \ref{FigDefectCount}.

\vspace{-0.1cm}
\subsection{Benchmarks}
Following the benchmarks on object or scene text detection datasets \cite{Pascal, Icdar2015}, average precision rate and F-mean are used for evaluation. A detection is correct only if the intersection of unit (IoU) between the detected bounding box and any of the ground truth box with the same class is larger than 0.33.

\vspace{-0.1cm}
\section{Approach}
\vspace{-0.2cm}
In this section, we describe the proposed in a specific way for detecting PCB defects from a pair of input images.


\subsection{Network Structure}
\begin{figure} [t]
\centering
\includegraphics[width=8.5cm]{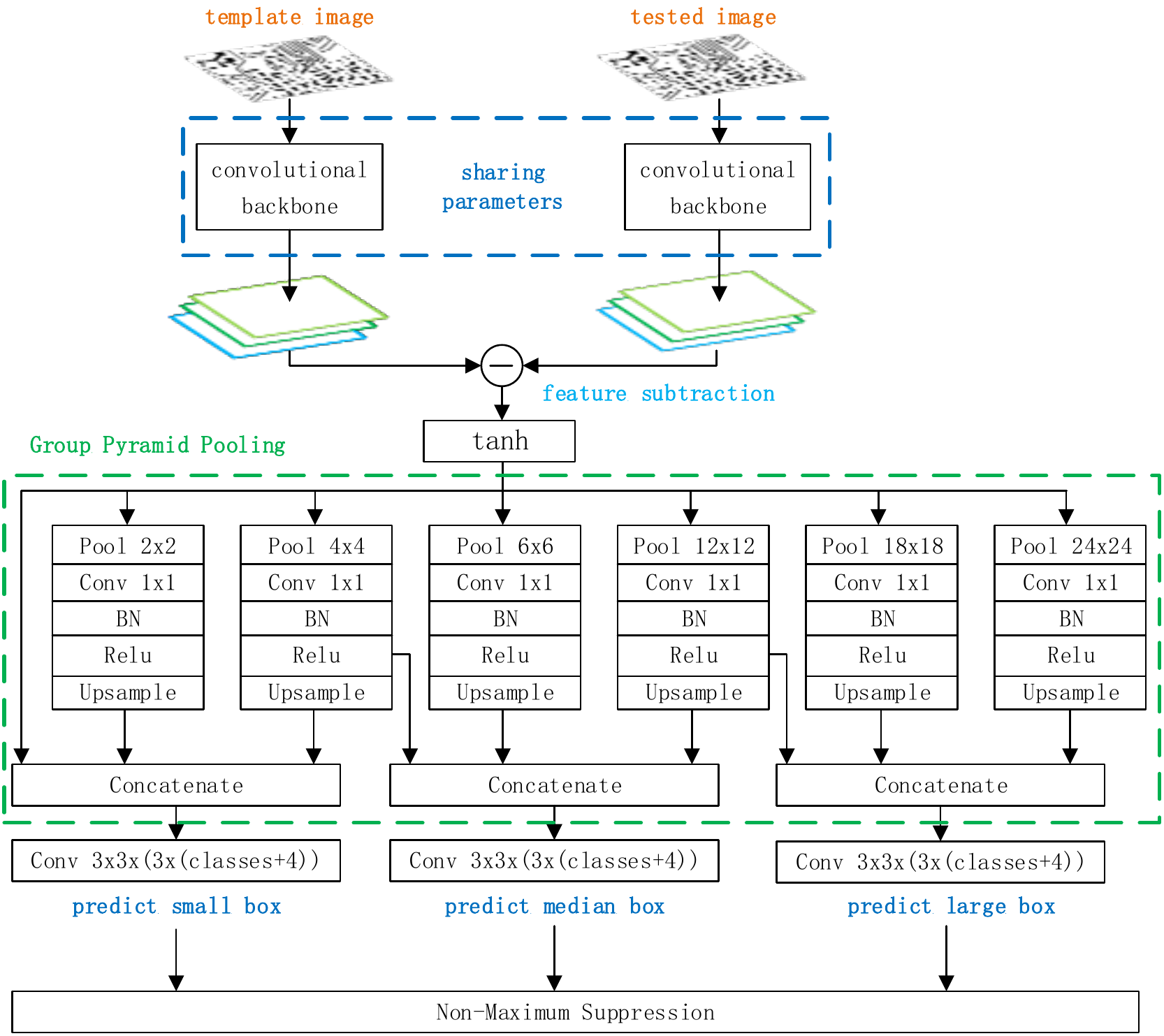}
\caption{An overview of the proposed model. The backbone can be any efficient convolutional model, e.g., VGG16-tiny \cite{VGG} or ResNet18 \cite{ResNet} without the last fully connected layers. We examine the average pooling and max pooling as the 'Pool' operation in GPP module in the experiments. 'BN' is abbreviation of batch normalization. Up-sample is implemented by bilinear interpolation and the target size is the same as the first input of each concatenated group. Each group in GPP module makes predictions in different scales.}
\label{FigStructure}
\vspace{-0.2cm}
\end{figure}

Instead of directly comparing the difference between the input image pair, a convolutional backbone with max pooling operation is first deployed for extracting features of translation and rotation invariance from the input images. Then, the differences of the features from template and tested images are calculated. A novel group pyramid pooling module is followed to obtain features in various resolutions. Similar to \cite{SSD, FPN}, we produce predictions of different scales from feature maps from the backbone. In Fig. \ref{FigStructure}, we show the overall structure of the proposed PCB defect detection model. 

\noindent {\bf{Group Pyramid Pooling (GPP) module}} Different from feature pyramid network (FPN) \cite{FPN} by merging features in different resolutions from coarse to fine, which increases the computational and storage cost, GPP module obtains features of various resolutions from a pyramid pooling structure. A similar pyramid structure has been studied in \cite{PSPNet}, however, GPP divides the pyramid pooling operations into group with overlaps to more precisely predict PCB defects of different scales. Actually, each group takes the responsibility for predicting the PCB defect from the default boxes \cite{SSD} in a specific scale, e.g. in the first group, it takes use of features from 1x1, 2x2 and 4x4 size of pooling to predict PCB defects with small bounding boxes. Besides, those groups share part of input features with its adjacent group to reduce the edge effect. 

\begin{table*}[h]
\caption{Evaluation result of mAP on DeepPCB dataset. 'AP' or 'MP' denotes the contrastive settings of average pooling or max-pooling operation in the GPP module.}
\label{TbAll}
\centering
\begin{tabular}{c|c|cccccc}
\hline
Method & mAP & open & short & mousebite & spur & copper & pin-hole \\
\hline
Image Processing \cite{PCBMorphology} & 89.3 & 88.2 & 87.6 & 90.3 & 88.9 & 91.5 & 89.2\\
\hline
SSD \cite{SSD} & 95.9 & 93.1 & 94.5 & 95.7 & 96.7 & 96.9 & 98.7 \\
YOLO \cite{YOLO} & 92.6 & 90.5 & 92.0 & 93.1 & 93.3 & 94.9 & 92.6\\
Faster \cite{Faster} & 97.6 & 96.8 & 95.4 & 97.9 & 98.7 & 97.4 & 99.5\\
\hline
ours-AP & 97.1 & 97.0 & 93.5 & 98.7 & 96.6 & 97.4 & \textbf{99.9} \\
ours-MP & \textbf{98.6} & \textbf{98.5} & \textbf{98.5} & \textbf{99.1} & \textbf{98.2} & \textbf{98.5} & 99.4\\
\hline
\end{tabular}
\end{table*}

\noindent {\bf{Prediction from convolutional feature maps}} Each output feature map from GPP module can produce a fixed set of detection predictions by several convolutional filters. As illustrated in the Fig. \ref{FigStructure}, the top convolutional layer output an $m \times n \times (3 \times ({\rm classes} + 4))$ map for prediction. Three kinds of default boxes \cite{SSD} with aspect ratio 0.5, 1.0, 2.0 are generated and center on each of $m \times n$ locations. The size of default boxes are configured manually as hyper-parameters, where 0.04, 0.08, 0.16 of the input image size (corresponding to the small, median, and large boxes in Fig. \ref{FigStructure}) are adopted in our setting. At each $m \times n$ location, it outputs the prediction of (i) classification: six types of PCB defects and one backgroud class, and (ii) localization: the translation offset between the centroids and the scaling ratio between the width and height  of the default boxes and the targets. Finally, non-maximum suppression (NMS) is applied to all the predictions from different scales to obtain the final prediction results.


\subsection{Objective Function}
Following the matching strategy in SSD \cite{SSD}, each ground truth box is first matched to the default box \cite{SSD} of the maximum jaccard overlap \cite{MultiBox}. Then, the default boxes with any ground truth box whose jaccard overlap is higher than 0.5 are matched. Those matched pairs can be described as: 
\begin{equation}
\mathcal{D}=\{(d, g)|{\rm jaccard\_overlap}(d,g)>0.5)\},
\end{equation}
 where $d=(d^{cx}, d^{cy}, d^{w}, d^{h})$ and $g=(g^{cx}, g^{cy}, g^{w}, g^{h})$ are the central point, width and height of default box and ground truth box respectively. Then, the objective function for box regression is defined as:
\begin{equation}
L_{\rm reg}=\sum_{(d_n, g_n)\in\mathcal{D}}\sum_{i\in{cx, cy, w, h}}{\rm smooth_{L1}}(l_n^i-t_n^i),
\end{equation}
where ${\rm smooth_{L1}(x)}=0.5x^2, {\rm if}~|x|<1 ~~{\rm or}~~ |x|-0.5$, otherwise. $l_n$ is the predicted offset between the default bounding box $d_n$ and the matched ground truth box $g_n$ and $t_n$ is the target offset, which is normalized by the default box size:
\begin{align*}
& t_n^j = (g_n^j-d_n^j)/d_n^j,\quad &j\in\{cx, cy\}, \\
& t_n^k = {\rm log}(\frac{g_n^k}{d_n^k}),\quad &k\in\{w, h\}.
\end{align*}
The classification loss for the type of PCB defect is calculated  by softmax loss, where we randomly select background default boxes to keep the ratio between the background (Bg) and foreground (Fg) bounding boxes at around 3 : 1.
\begin{equation}
L_{\rm cls}=-\sum_{d_n\in {\rm Fg}}{\rm log}(c_n^p) - \sum_{d_n\in {\rm Bg}}{\rm log}(c_n^0),
\end{equation}
where $c_n^p$ is the predicted probability that the target in the box $d_n$ belongs to class $p$. Notice that the class index for background box is set to 0.

\section{Experiments}
In this section, extensive experiments are carried out to evaluate the proposed model as well as some advanced object detection model  for PCB defect detection on DeepPCB dataset. As a competitive method, \cite{PCBMorphology} uses image processing techniques based on the pair of template and tested image to detect and distinguish PCB defects. Since models \cite{SSD, YOLO, Faster} based on deep neural network show their great power in object detection tasks, we also examine those methods' abilities for PCB defect detection with slight modifies: (i) a convolutional backbone, e.g., VGG-tiny \cite{VGG} is first applied to efficiently extract features of both the template and tested image; (ii) feature subtraction is adopted to merge the last feature maps of the backbone from the input image pair. 

We train our model on a single Titan X GPU using Adam with initial learning rate $10^{-3}$, 0.0005 weight decay, 500 epochs and batch size 16. The learning rate decays 0.33 every 100 epoch. The whole training process takes about 0.5 day.

\begin{figure}[h]
\centering
\includegraphics[width=8cm]{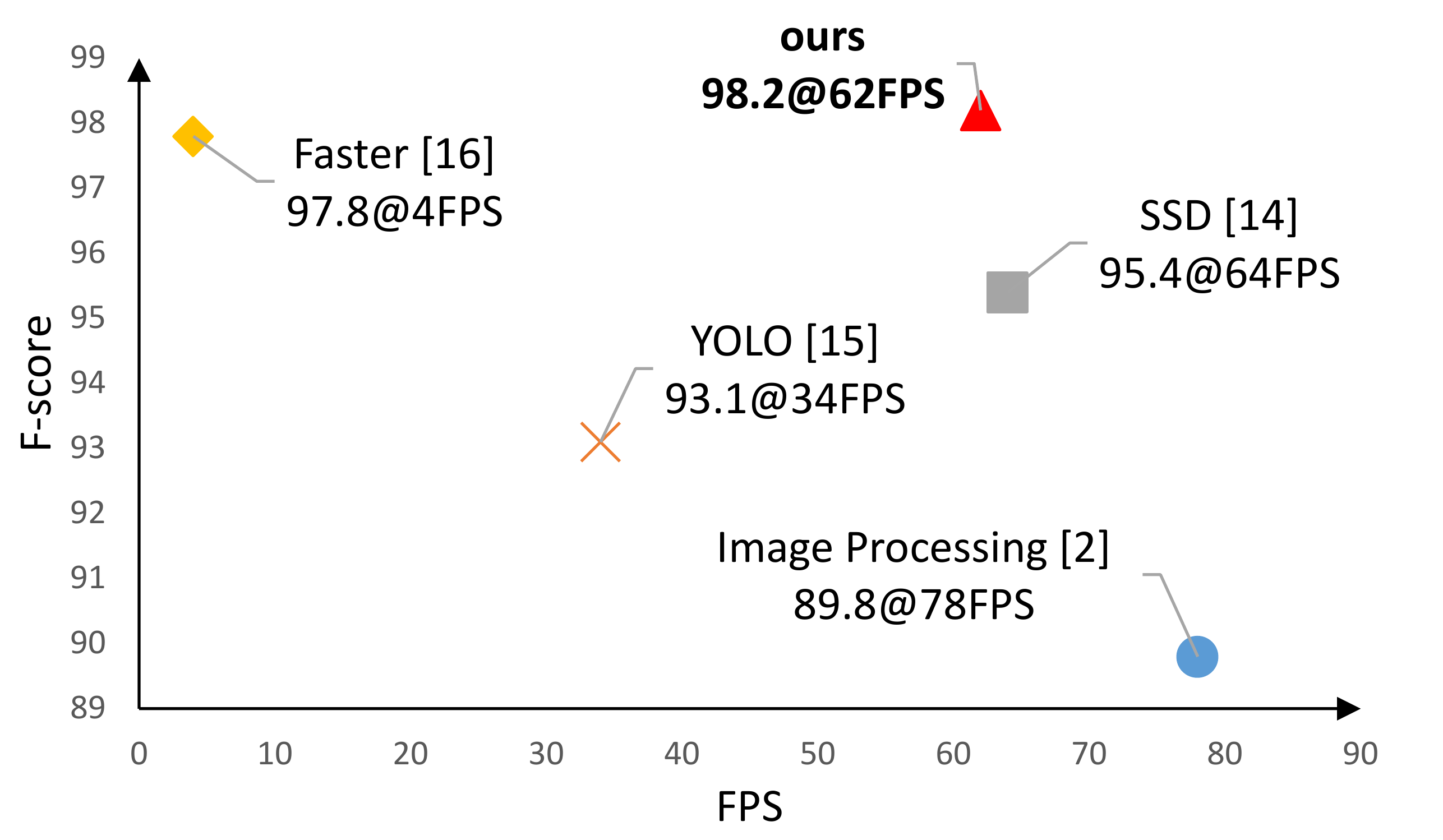}
\caption{Comparison of performance and detection speed of the PCB defect detection methods on DeepPCB dataset.}
\label{FigEfficiency}
\vspace{-0.2cm}
\end{figure}

\vspace{-0.15cm}
\subsection{Results on DeepPCB Dataset}
This section provides quantitative evaluations for various methods of PCB defect detection on DeepPCB dataset. For data argumentation of the models based on deep neural networks, the template and tested image of each pair are simultaneously randomly horizontal/vertical flipped with probability of 0.5 and then are randomly cropped into size of 512 x 512. In Table \ref{TbAll}, we illustrate the evaluation result on DeepPCB dataset. For both one-stage models \cite{SSD, YOLO} and more complecated two-stage model \cite{Faster}, as well as the algorithm based on image processing techniques \cite{PCBMorphology}, the proposed model improves mean average precision from $1.0\%$ to $9.3\%$. Experiment results also show that the max pooling in GPP module improves $1.5\%$ mAP than the average pooling. Although \cite{Faster} achieves competitive performance, however, it takes much longer time for inference. In Fig. \ref{FigEfficiency}, we summarize the inference speed measured by frame per second (FPS) and the F-score of the comparative models for PCB defect detection.

\vspace{-0.2cm}
\begin{table}[h]
\caption{Ablation experiment results on Deep PCB dataset.}
\label{TbAblation}
\centering
\begin{tabular}{c|ccc}
\hline
Method &  precision & recall & F-mean\\
\hline
SSD-FPN & 94.9 & 96.8 & 95.8\\
ours-non-GPP & 94.3 & 96.3 & 95.3\\
ours-MP & \textbf{98.2} & \textbf{98.1} & \textbf{98.2} \\
\hline
\end{tabular}
\vspace{-0.4cm}
\end{table}

\subsection{Ablation Study on Group Pyramid Pooling Module}
To validate the effectiveness of group pyramid pooling module for PCB defect detection, we carry out experiments with a compartive setting: only keeping the 1x1, 4x4 and 12x12 pooling operations (denoted as ours-non-GPP) and removing other pooling operations in GPP module in Fig. \ref{FigStructure}. Besides, we also compare to combine SSD model with feature pyramid network (FPN) \cite{FPN} to merge features in various resolutions for prediction, which shares similar structure as in DSSD \cite{DSSD} (denoted as SSD-FPN). Again, all the models are established by the same convolutional backbone: VGG-tiny \cite{VGG}. In Table \ref{TbAblation}. the evaluation result shows that the F-score of the proposed model surpasses the comparative model without GPP module by $3.9\%$ and the SSD-FPN structure by $3.3\%$.

\section{Conclusion}
This work contributes DeepPCB, a large-scale PCB dataset for PCB defect detection with annotations of the positions and six common types of PCB defects. To the best of our knowledge, this is the first dataset in terms of the scales and precise annotations. A novel deep module is proposed, namely group pyramid pooling, that efficiently combines features in different resolutions and makes predictions for detecting PCB defects in various scales. Through extensive experiment, we demonstrate that the proposed architecture with GPP module can achieve state-of-the-art performance while consuming very low computational time. The public DeepPCB dataset will further facilitate future works.

\vfill\pagebreak

\small
\bibliographystyle{IEEEbib}
\bibliography{tang}

\end{document}